\documentclass[sigconf]{acmart}
\AtBeginDocument{%
  }

\copyrightyear{2026}
\acmYear{2026}
\setcopyright{cc}
\setcctype{by}
\acmConference[WWW '26]{Proceedings of the ACM Web Conference 2026}{April 13--17, 2026}{Dubai, United Arab Emirates}
\acmBooktitle{Proceedings of the ACM Web Conference 2026 (WWW '26), April 13--17, 2026, Dubai, United Arab Emirates}
\acmPrice{}
\acmDOI{10.1145/3774904.3792900}
\acmISBN{979-8-4007-2307-0/2026/04}

\usepackage{multirow}
\usepackage{graphicx} % Add this line in the preamble

\begin{document}

\title{Is More Context Always Better? Examining LLM Reasoning Capability for Time Interval Prediction}


% \author{Anonymous Author(s)}
% \date{}
% \settopmatter{authorsperrow=4}

\author{Yanan Cao}
\authornote{Highlighted authors contributed equally to this research.}
\affiliation{%
  \institution{Walmart Global Tech}
  \city{Sunnyvale}
  \state{CA}
  \country{USA}
}
\email{yanan.cao@walmart.com}

\author{Farnaz Fallahi}
\authornotemark[1]
\affiliation{%
  \institution{Walmart Global Tech}
  \city{Sunnyvale}
  \state{CA}
  \country{USA}
}
\email{farnaz.fallahi@walmart.com}

\author{Murali Mohana Krishna Dandu}
\authornotemark[1]
\affiliation{%
  \institution{Walmart Global Tech}
  \city{Sunnyvale}
  \state{CA}
  \country{USA}
}
\email{murali.dandu@walmart.com}

\author{Lalitesh Morishetti}
\authornotemark[1]
\affiliation{%
  \institution{Walmart Global Tech}
  \city{Sunnyvale}
  \state{CA}
  \country{USA}
}
\email{lalitesh.morishetti@walmart.com}

\author{Kai Zhao}
\authornote{Work done while at Walmart Global Tech.}
\affiliation{%
  \institution{Walmart Global Tech}
  \city{Sunnyvale}
  \state{CA}
  \country{USA}
}
\email{kaizhaofrank@gmail.com}

\author{Luyi Ma}
\affiliation{%
  \institution{Walmart Global Tech}
  \city{Sunnyvale}
  \state{CA}
  \country{USA}
}
\email{luyi.ma@walmart.com}

\author{Sinduja Subramaniam}
\affiliation{%
  \institution{Walmart Global Tech}
  \city{Sunnyvale}
  \state{CA}
  \country{USA}
}
\email{sinduja.subramaniam@walmart.com}

\author{Jianpeng Xu}
\affiliation{%
  \institution{Walmart Global Tech}
  \city{Sunnyvale}
  \state{CA}
  \country{USA}
}
\email{jianpeng.xu@walmart.com}

\author{Evren Korpeoglu}
\affiliation{%
  \institution{Walmart Global Tech}
  \city{Sunnyvale}
  \state{CA}
  \country{USA}
}
\email{ekorpeoglu@walmart.com}

\author{Kaushiki Nag}
\affiliation{%
  \institution{Walmart Global Tech}
  \city{Sunnyvale}
  \state{CA}
  \country{USA}
}
\email{kaushiki.nag@walmart.com}

\author{Sushant Kumar}
\affiliation{%
  \institution{Walmart Global Tech}
  \city{Sunnyvale}
  \state{CA}
  \country{USA}
}
\email{sushant.kumar@walmart.com}

\author{Kannan Achan}
\affiliation{%
  \institution{Walmart Global Tech}
  \city{Sunnyvale}
  \state{CA}
  \country{USA}
}
\email{kannan.achan@walmart.com}

\renewcommand{\shortauthors}{Yanan Cao et al.}

\begin{abstract}

% \YC{The anwer of the title is now proper level of context works the best}

% \LuM{example of how to use comment, this is for Luyi} \KZ{this is for Kai} \FF{this is for Farnaz} \MD{this is for Murali} \LMo{this is for Lalitech}
Large Language Models (LLMs) have demonstrated impressive capabilities in reasoning and prediction across different domains. Yet, their ability to infer temporal regularities from structured behavioral data remains underexplored. This paper presents a systematic study investigating whether LLMs can predict time intervals between recurring user actions, such as repeated purchases, and how different levels of contextual information shape their predictive behavior. Using a simple but representative repurchase scenario, we benchmark state-of-the-art LLMs in zero-shot settings against both statistical and machine-learning models. Two key findings emerge. First, while LLMs surpass lightweight statistical baselines, they consistently underperform dedicated machine-learning models, showing their limited ability to capture quantitative temporal structure. Second, although moderate context can improve LLM accuracy, adding further user-level detail degrades performance. These results challenge the assumption that “more context leads to better reasoning.” Our study highlights fundamental limitations of today’s LLMs in structured temporal inference and offers guidance for designing future context-aware hybrid models that integrate statistical precision with linguistic flexibility.

\end{abstract}





\maketitle

\section{Introduction}

Predicting the time until a user's next action is a fundamental yet often overlooked dimension of sequential behavioral modeling on the Web \cite{cao2025s2srec2, ma2025grace}. This interval, whether for replenishing groceries or renewing a subscription, tells us about users' habits and needs. Accurately forecasting this interval is therefore a key component of creating personalized experiences. Traditional machine learning (ML) and statistical approaches have long been used to estimate these intervals with structured data. However, the recent emergence of LLMs raises a question: can an LLM reason about time intervals as effectively as specialized predictive models? 

Though LLMs show strong capability in language understanding, mathematical reasoning and multi-step logical inference\, their ability in structured temporal reasoning remains largely unexamined. Most studies focus on what LLMs can infer from text \cite{zhang2025no}, not on whether they can capture quantitative regularities hidden in behavioral data. 

\begin{figure}[!htbp]
    \centering
    \includegraphics[width=1\linewidth]{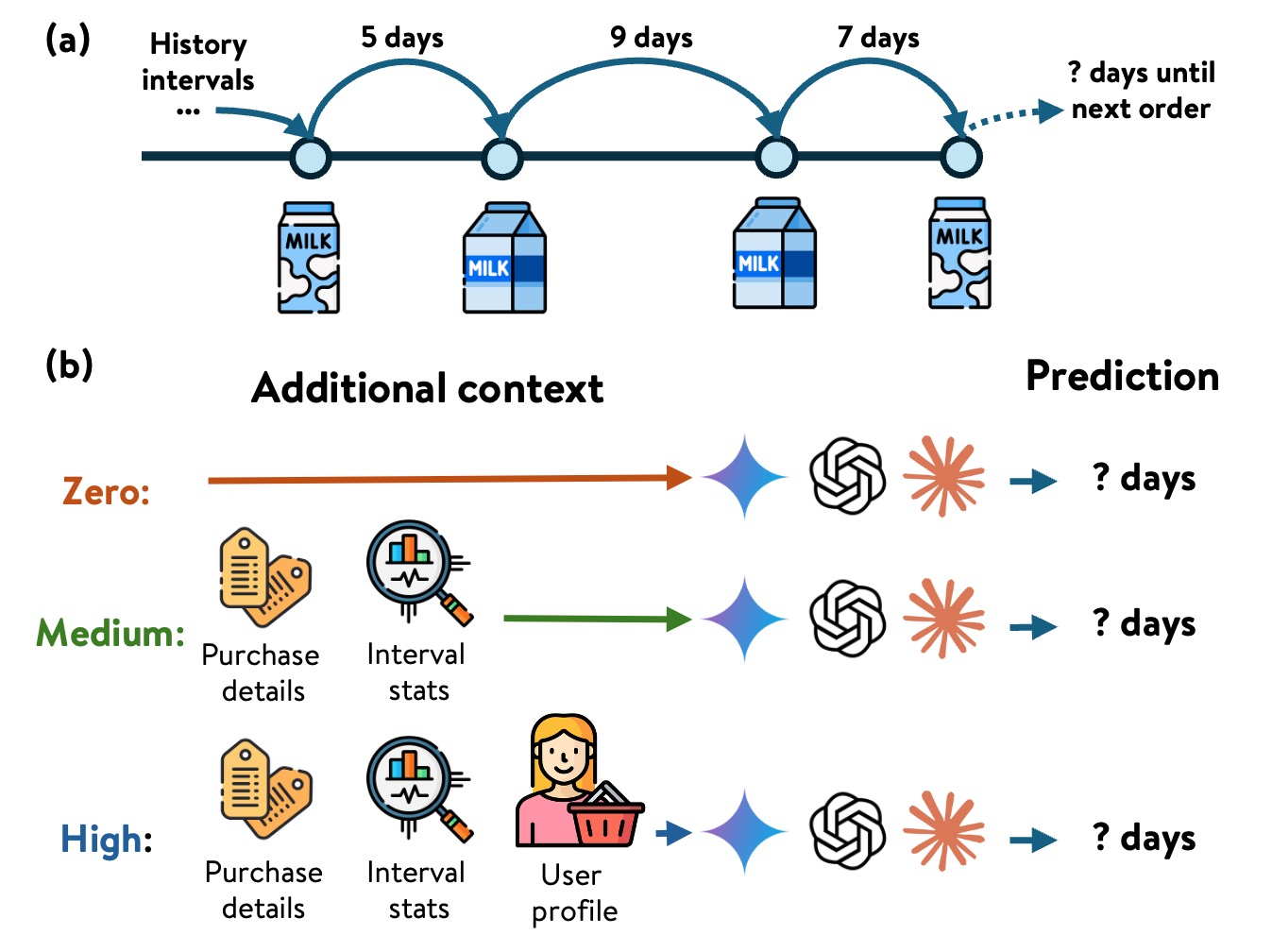}
    \caption{Illustration of the interval-prediction task and the three prompting conditions. The example is about repeated milk purchases with varying intervals (5 days → 9 days → 7 days → 6 days). The model observes historical intervals for a product category and predicts the next interval under zero, medium, and high context settings.}
    \label{fig:illustration}
\end{figure}

Recent work applies LLMs to temporal forecasting by converting time series into textual sequences. Models such as LLMTime \cite{gruver2024largelanguagemodelszeroshot} treats numerical time series as digit string tokens, enabling zero-shot forecasting without model modifications. TIME-LLM \cite{jin2024timellmtimeseriesforecasting} reprograms LLM by learning an alignment layer that translates time series values into its lexical space. Methods like LSTPrompt \cite{liu2024lstpromptlargelanguagemodels} prompt LLMs to reason separately about long and short range patterns. Despite these advances, their benefits over classical models remain uncertain. \cite{tang2024timeseriesforecastingllms} shows that LLMs can handle periodic patterns but struggle with irregular ones, although external context can help. In user modeling, \cite{du2025justwhatwhenintegrating}\cite{dong2025see} incorporate irregular time intervals into LLMs for sequential recommendation, yet the task remains focused on what to recommend rather than when the event occurs.

In contrast to prior work on continuous time-series forecasting, our study focuses on discrete, event-based inter-purchase intervals in web behavior modeling. The prevailing literature often assumes that richer contextual data will inherently improve LLM reasoning, yet this has rarely been tested for quantitative, time-dependent tasks. We directly examine whether supplying LLMs with semantic cues (e.g., product attributes, quantities, day-of-week patterns) enables better timing predictions than relying on numerical histories alone. To our knowledge, this hypothesis has not yet been systematically validated in this specific domain. A simple example in Figure \ref{fig:illustration} illustrates the task: a household repeatedly purchases milk with gaps of 5, 7, and 9 days, and the model must infer the next interval. This small-scale scenario captures a broad challenge central to web intelligence: inferring behavioral rhythms from sparse, event-driven user interactions.

In this work, we conduct the first systematic study comparing ChatGPT \cite{achiam2023gpt}, Claude \cite{anthropic2024claude3}, and Gemini \cite{comanici2025gemini} with both statistical and ML baselines on this time-interval prediction tasks. We design progressively richer prompts (in zero-shot settings), from zero-context to medium and high-context scenarios incorporating behavioral summaries and average consumption patterns. Our work addresses two questions:

\noindent\textbf{RQ1:} Can LLMs outperform traditional machine learning models in inter-purchase interval prediction?\\
\textbf{RQ2:} Does providing richer contextual information improve LLM performance on temporal interval reasoning tasks?

Across all configurations, LLMs outperform simple statistical estimators yet fall short of traditional machine-learning models such as XGBoost on quantitative accuracy metrics. Although medium-level context can yield small gains, additional high-level narrative or behavioral context often reduces accuracy, suggesting that richer prompts may introduce noise that distracts LLMs from core numerical patterns. These findings provide new empirical evidence on the limits of LLM reasoning in structured temporal tasks and  motivates deeper investigation into how LLMs internalize temporal signals.

\section{Experimental Setting}

\subsection{Datasets}

We evaluate our approach on two real-world e-commerce datasets: proprietary grocery data and the public Instacart dataset \cite{yasserh_instacart_2017}. In both datasets, user-category pairs with fewer than five purchases are excluded, and intervals longer than 20 days are removed. The proprietary dataset contains 5{,}780 users across 12 product types (1.03 types/user average) with over 110{,}000 purchase records. The Instacart dataset includes 2{,}661 users across 10 categories (1.87 categories/user average) with roughly 194{,}000 orders. Both datasets span diverse categories (Fresh Produce, Dairy, Packaged Goods, Snacks) capturing variability in purchase periodicity and seasonality for temporal reasoning comparison.

\subsection{Models}
We evaluate three categories of models for temporal interval prediction: LLMs, ML models, and statistical estimator baselines.

\textbf{LLM Models.}
We evaluate several state-of-the-art LLMs to examine their ability to predict temporal purchase intervals from structured behavioral data. Specifically, we consider GPT-4o, Gemini-2.5 Pro, and Claude 3.5 Sonnet, prompted under systematically varied context levels to study how contextual richness influences temporal reasoning. As illustrated in Figure 2, each model receives a textual description of a user’s past purchase events and is asked to predict the next inter-purchase interval in days. To control the amount of information provided, we design three prompting levels, allowing us to test how much information an LLM needs for accurate timing predictions. Zero-context isolates pure numeric reasoning from intervals alone. Medium-context adds product descriptions and basic statistics to see whether product metadata or simple statistics help. High-context includes richer behavioral signals to evaluate whether LLMs benefit from features commonly used in traditional forecasting. This progression provides a controlled way to assess the impact of different information levels. All model outputs are parsed into numeric predictions and evaluated under a unified scoring framework. We focused on zero-shot (instead of few-shot) evaluation as our is goal is to mainly assess LLM's inherent temporal reasoning capabilities without task-specific guidance, leaving advanced prompt tuning strategies to future work.

\begin{figure}[!htbp]
    \centering
    \includegraphics[width=1\linewidth]{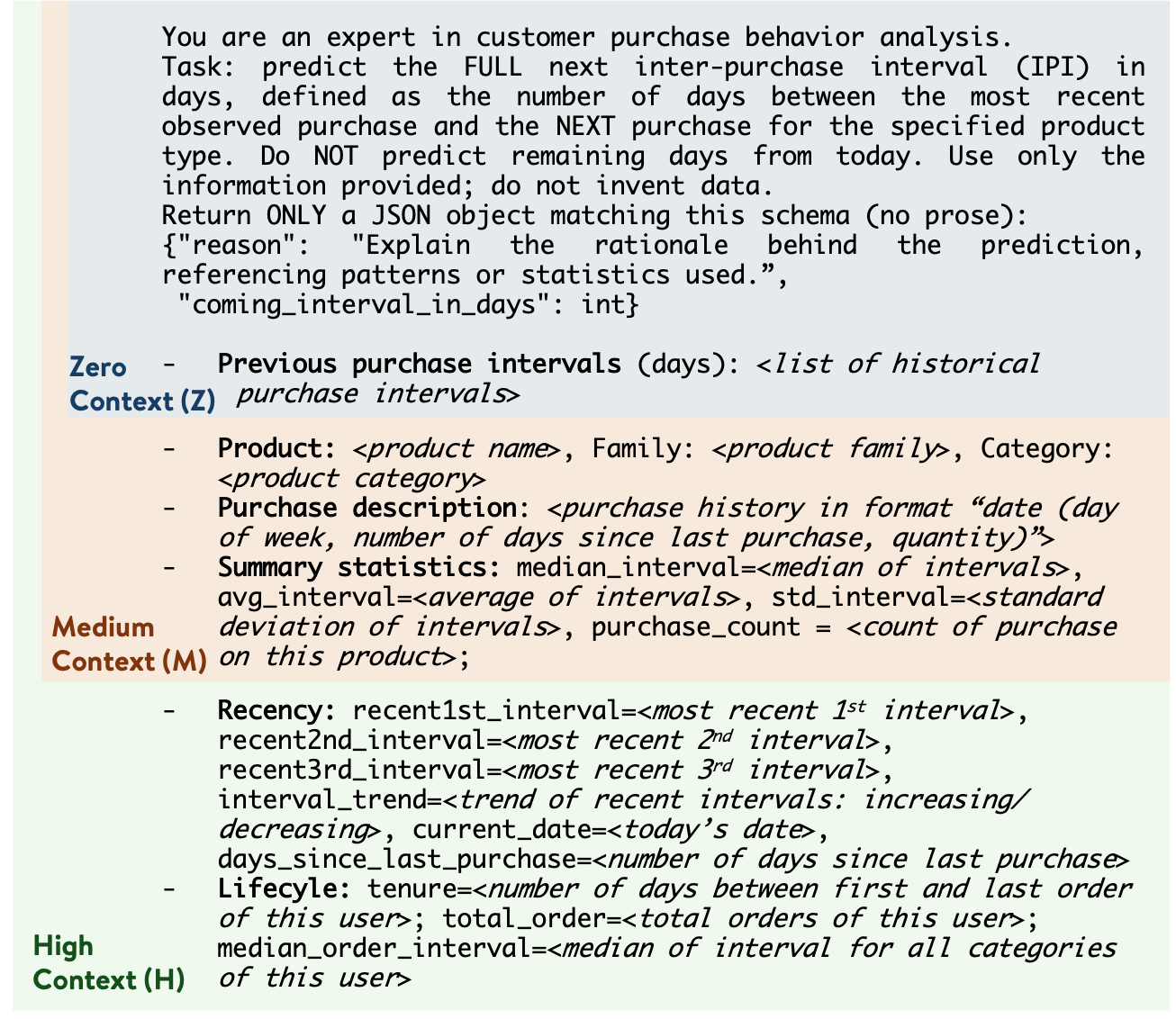}
    \caption{Prompt designs for three context levels: Zero (historical intervals only), Medium (product metadata, summary statistics), and High (recency features, user lifecycle information).}
    \label{fig:prompts}
\end{figure}

%Prompt designs for three context levels: Zero (intervals only), Medium (product metadata, summary statistics), and High (recency features, user lifecycle information).

%Overview of the prompt designs used for evaluating LLM reasoning under three context levels: Zero Context (historical intervals only), Medium Context (product metadata and basic summary statistics), and High Context (recency features and user-level lifecycle information).

\textbf{ML Models.}
To establish structured-data baselines, we implement classical ML models that learn temporal dependencies from numerical features, such as historical inter-purchase intervals, user recency and category-level purchase frequency. We trained three representative machine learning models in industry that are commonly used (RandomForest regression, XGBoost regression, and Multi-layer deep neural network), optimized using median quantile loss to better accommodate the discrete distribution of the target variable and align with business metrics for next-purchase prediction in practice. Our experiments demonstrate that models trained with quantile loss achieves superior performance across both datasets. This structured setup reflects the dominant paradigm in tabular and demand-forecasting systems, providing a grounded benchmark for comparison against LLM-based interval reasoning.

\textbf{Statistical Models.}
We further include lightweight statistical estimators that capture baseline temporal regularities without learning. These include the mean, median, and exponential-moving-average (EMA) of previous intervals, each computed at user and product type level. Such models are widely used in retail analytics and serve as interpretable lower-bound baselines that reveal how much improvement more sophisticated learners or reasoning-based systems can realistically achieve.

\section{Experimental Results}
The results of our experiments provide a clear empirical view of how different modeling paradigms perform on structured temporal reasoning. We compare multiple metrics across all models, examine the effects of varying context on LLM prediction behavior, and analyze representative cases where LLM reasoning diverges from data-driven estimates. 

\subsection{Main Results}

\begin{table}[!htbp]
\centering
\caption{Accuracy and error metrics across context levels, with best ML and statistical baselines. Bold values indicate the overall best performance across all models. Underlined values indicate the best among LLMs.}
\label{tab:llm_combined_metrics_onecol}
\setlength{\tabcolsep}{2pt}
\renewcommand{\arraystretch}{0.92}
\begin{tabular}{lcccccc}
\toprule
\multicolumn{1}{c}{\multirow{2}{*}{\textbf{Model}}} &
\multicolumn{3}{c}{\textbf{Accuracy Metrics ↑}} &
\multicolumn{3}{c}{\textbf{Error Metrics ↓}} \\
\cmidrule(lr){2-4}\cmidrule(lr){5-7}
& \textbf{TA@0} & \textbf{TA@1} & \textbf{TA@2} &
\textbf{RMSE} & \textbf{MAE} & \textbf{MAPE} \\
\midrule
\multicolumn{7}{c}{\textbf{Proprietary data}} \\
\midrule
GPT-4o-Z    & 5.75 & 12.72 & 18.65 & 23.66 & 15.50 & 73.03 \\
GPT-4o-M   & 6.13 & 13.83 & 19.92 & 22.95 & 14.76 & 66.78 \\
GPT-4o-H   & 5.32 & 12.72 & 18.48 & 24.83 & 16.39 & 76.59 \\
\midrule
Gemini-2.5-Z  & 6.38 & 13.57 & 19.68 & 23.44 & 15.17 & 63.79 \\
Gemini-2.5-M  & 6.15 & 13.95 & 19.72 & 23.91 & 15.39 & 67.79 \\
Gemini-2.5-H  & 6.20 & 13.42 & 19.25 & 24.20 & 15.66 & 71.38 \\
\midrule
Claude-3.5-Z & 5.98 & 13.50 & 19.72 & 22.26 & 14.17 & 64.27 \\
Claude-3.5-M & 6.55 & 14.58 & \underline{20.97} & \underline{21.93} & \underline{13.85} & \underline{57.45} \\
Claude-3.5-H & \underline{6.75} & \underline{14.63} & 20.95 & 22.11 & 14.11 & 59.61 \\
\midrule
ML Best      & \textbf{9.48} & \textbf{22.98} & \textbf{33.93} & \textbf{9.97} & \textbf{7.18} & \textbf{29.92} \\
Stat Best    & 4.42 & 13.15 & 20.32 & 22.46 & 14.25 & 55.41 \\
\midrule
\multicolumn{7}{c}{\textbf{Instacart data}} \\
\midrule
GPT-4o-Z   & 6.54 & 15.46 & 22.16 & 30.11 & 16.13 & 77.39 \\
GPT-4o-M   & \underline{7.32} & 16.04 & 22.98 & 28.56 & 15.09 & 66.80 \\
GPT-4o-H   & 6.00 & 14.12 & 20.12 & 31.05 & 17.01 & 84.03 \\
\midrule
Gemini-2.5-Z & 7.30 & 15.76 & 22.82 & 28.85 & 15.27 & 64.13 \\
Gemini-2.5-M & 7.28 & \underline{16.64} & \underline{23.06} & 28.36 & 15.05 & \underline{59.43} \\
Gemini-2.5-H & 6.26 & 14.80 & 21.36 & 29.46 & 16.17 & 74.38 \\
\midrule
Claude-3.5-Z & 6.22 & 14.48 & 22.20 & \underline{26.88} & 14.18 & 67.71 \\
Claude-3.5-M & 6.02 & 14.24 & 21.82 & 26.92 & \underline{13.93} & 62.29 \\
Claude-3.5-H & 6.92 & 15.10 & 22.44 & 27.50 & 14.42 & 64.31 \\
\midrule
ML Best      & \textbf{8.46} & \textbf{22.62} & \textbf{33.42} & \textbf{9.17} & \textbf{6.55} & \textbf{35.04} \\
Stat Best    & 5.90 & 15.34 & 23.00 & 27.97 & 14.52 & 56.34 \\
\bottomrule
\end{tabular}
\end{table}

We assess model performance using three categories of metrics: (1) standard regression metrics including Root Mean Squared Error (RMSE), Mean Absolute Error (MAE), and Mean Absolute Percentage Error (MAPE). (2) business-oriented accuracy metrics, defined as Tolerance Accuracy (TA@k) which is the proportion of predictions where the absolute error is $k$ days or less (i.e., $|y - \hat{y}| \le k$). This metric directly measures the model's utility in a real-world setting where exact precision is not always required. We report TA@0 (exact-day accuracy), TA@1 (±1-day accuracy), and TA@2 (±2-day accuracy). (3) Deployment performance is measured by the cost and latency required to generate a prediction.
 % and the coefficient of determination ($R^2$)

\textbf{RQ1: LLMs vs. Traditional ML}
A primary finding from \ref{tab:llm_combined_metrics_onecol}, consistent across both the proprietary and Instacart datasets, is the ML model's dominant performance. It achieves over 50\% improvement on most key metrics compared to the best-performing LLM. For instance, on the proprietary dataset, the ML model's MAPE of 29.92\% is 92.0\% better than the best-performing LLM (Claude-3.5 Medium at 57.45\%). On the business-critical "TA@1" metric, the ML model scores 22.98\%, beating the best LLM's 14.63\% (Claude-3.5 High) by 57.1\%. This performance gap is likely due to a fundamental "impedance mismatch" for the LLM. The ML model directly operates on structured numerical features, whereas the LLM must first translate qualitative linguistic descriptions into an internal representation before performing regression, introducing significant error. Notably, the disparity in performance is significantly larger for error metrics than for accuracy metrics, as LLMs' performance on TA@k is competitive. This suggests that while LLMs fail at quantitative precision (i.e. pinpointing the exact day), they are better at approximate temporal classification. This aligns with the business goal for replenishment, where knowing a user will buy "in the next 48 hours" is often sufficient.

Another notable observation is that LLMs outperform the “stat best” median baseline across both datasets and metrics, indicating that they extract additional signal from contextual descriptions beyond simple historical central tendency. This suggests that LLMs incorporate contextual cues rather than merely reproducing medians, yielding more informative estimates, especially for approximation rather than exact-day prediction. These results indicate that LLMs have useful but limited temporal reasoning capabilities.

\begin{figure}[!htbp]
    \centering    
    \includegraphics[width=1\linewidth]{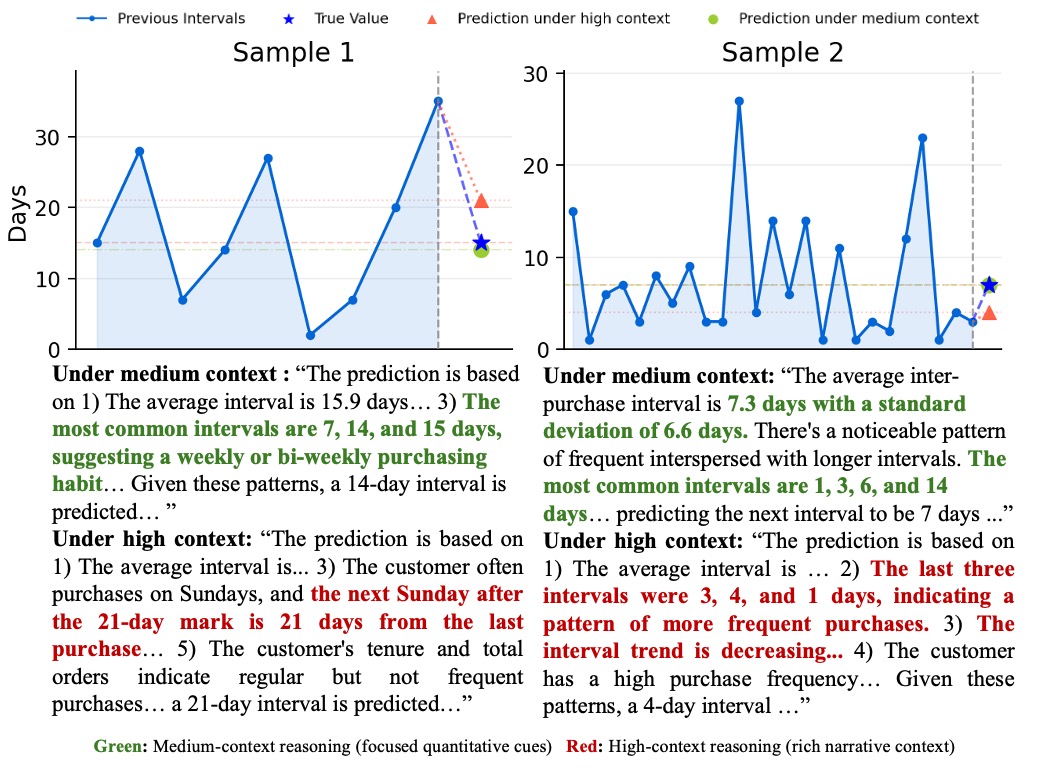}
    \caption{Examples of Claude-3.5-Sonnet prediction and reasoning with different levels of context.}
    \label{fig:casestudy}
\end{figure}

\textbf{RQ2: The Impact of Context on LLM Performance} Our second research question examines how different levels of context affect LLM performance. Across both datasets, medium-level context, including basic product attributes, simple statistics, and concise user history, consistently improves both model metrics and business-oriented accuracy. In contrast, high-level narrative context often degrades performance, sometimes matching the zero-context baseline. This suggests a context-as-noise effect: focused, decision-relevant information aids temporal reasoning, whereas semantically rich but temporally irrelevant details distract the model. As illustrated in Figure \ref{fig:casestudy}, green highlights show that medium-context prompts encourage the model to rely on stable quantitative cues, producing correct or near-correct predictions, while red highlights show that high-context prompts surface narrative details that models overweight, diverting attention from the underlying temporal structure and leading to erroneous predictions.

Across LLM families, Claude-3.5 performs most consistently and achieves the best results on the internal dataset. On the Instacart dataset, Claude-3.5 and Gemini-2.5 achieve top performance on different metrics. In terms of efficiency, GPT-4o is the fastest and cheapest (around 1.2--1.4s latency, >\$0.003 per call), while Claude-3.5 is moderately slower (around 2--4s) and 3$\times$ more expensive, and Gemini-2.5 has the highest latency (around 15--19s) with lower overall accuracy.

Our study presents a framework for time-interval prediction, establishes strong baselines, and positions LLMs between statistical and machine-learning models. We show that targeted and compact context can aid LLM reasoning, while excessive narrative detail can obscure the temporal signal. These findings highlight both the potential and limitations of LLMs for structured temporal inference and point toward hybrid models combining statistical precision with linguistic flexibility.

% \clearpage
\bibliographystyle{ACM-Reference-Format}
\bibliography{bibfile}

\end{document}